\documentclass[preprint]{elsarticle}
\usepackage[hyphens]{url}  

\usepackage{hyperref}

\journal{Reliability Engineering and System Safety}

\usepackage{amsmath,amssymb,amsfonts}
\usepackage{subcaption}
\usepackage{algorithmic}
\usepackage{graphicx}
\usepackage{ulem}
\usepackage{acronym}
\usepackage[utf8]{inputenc} 
\usepackage[T1]{fontenc}    
\usepackage{url}            
\usepackage{booktabs}       
\usepackage{nicefrac}       
\usepackage{microtype}      
\usepackage{bm}
\usepackage{adjustbox}
\usepackage{multirow}
\usepackage{mathtools}
\usepackage{subcaption}
\usepackage{lineno}
\usepackage{setspace}
\usepackage{enumitem}
\usepackage{comment}
\usepackage{array, makecell} %
\usepackage{bbm}
\usepackage[table]{xcolor}
\usepackage{caption}
\usepackage{booktabs}

\usepackage{algorithm2e}
\usepackage{subcaption}

\usepackage{ulem}
\usepackage{xcolor}

\acrodef{0}[0]{Left-Hand Side}
\acrodef{1}[1]{Right-Hand Side}
\acrodef{dl}[DL]{Deep Learning}
\acrodef{ml}[ML]{Machine Learning}
\acrodef{cnn}[CNN]{Convolutional Neural Network}
\acrodef{nn}[NN]{Neural Network}
\acrodef{ann}[ANN]{Artificial Neural Networks}
\acrodef{mlp}[MLP]{Multilayer Perceptron}
\acrodef{rbf}[RBF]{Radial Basis Function}
\acrodef{lstm}[LSTM]{Long-Short Term Memory}
\acrodef{rnn}[RNN]{Recurrent Neural Network}
\acrodef{auc}[AUROC]{Area Under the Curve Receiver Operating Characteristics}

\RestyleAlgo{ruled}

\begin{document}

\begin{frontmatter}

\title{Detecting train driveshaft damages using accelerometer signals and Differential Convolutional Neural Networks}

\author[mymainaddress]{Antía López Galdo\corref{eq}}

\author[mymainaddress,secondaryaddress]{Alejandro Guerrero-López\corref{eq}}\ead{alexjorguer@tsc.uc3m.es}

\author[mymainaddress,secondaryaddress]{Pablo M. Olmos}

\author[mecadress]{María Jesús Gómez García}


\cortext[eq]{Equally contributed}

\address[mymainaddress]{Department of Signal Processing and Communications, Universidad Carlos III de Madrid, Leganés, 28911, Spain}
\address[secondaryaddress]{Instituto de Investigación Sanitaria Gregorio Marañón, Madrid, 28007, Spain}
\address[mecadress]{Department of Mechanical Engineering, Universidad Carlos III de Madrid, Leganes, 28911, Spain}

\begin{abstract}
Railway axle maintenance is critical to avoid catastrophic failures. Nowadays, condition monitoring techniques are becoming more prominent in the industry to prevent enormous costs and damage to human lives. 

This paper proposes the development of a railway axle condition monitoring system based on advanced 2D-\ac{cnn} architectures applied to time-frequency representations of vibration signals. For this purpose, several preprocessing steps and different types of \ac{dl} and \ac{ml} architectures are discussed to design an accurate classification system. The resultant system converts the railway axle vibration signals into time-frequency domain representations, i.e., spectrograms, and, thus, trains a two-dimensional \ac{cnn} to classify them depending on their cracks. The results showed that the proposed approach outperforms several alternative methods tested. The \acs{cnn} architecture has been tested in 3 different wheelset assemblies, achieving AUC scores of 0.93, 0.86, and 0.75 outperforming any other architecture and showing a high level of reliability when classifying 4 different levels of defects. 

\end{abstract}


\end{frontmatter}


\section{Introduction}

In industry, the development of new technologies and the trend towards process automation have opened up new lines of research, such as maintenance prediction \cite{mourtzis_vlachou_2018, nguyen_medjaher_2019} or estimation of remaining bearing life \cite{li_zhang_ding_2019, ding_jia_miao_huang_2021}. In terms of maintenance, condition-based maintenance is essential in industrial processes to achieve high system safety, reliability, and availability \cite{advances,lu_sun_zhang_feng_kang_fu_2018}.
 
Condition monitoring aims to continuously monitor and report on the condition of a machine during operation. Knowing the standard operating patterns is crucial to detect disturbances and take the appropriate action; therefore, recent studies are focused on this field \cite{yao2016pitch,pan2019incipient, liu2020vibration,zhang2022piezoelectric}. This type of maintenance implies that the machine is stopped only when necessary, thus reducing costs and time and increasing productivity. 

The dynamic behaviour of rotating machines in the presence of a defect has been studied in detail in recent decades. More specifically, with respect to railway axles, the literature \cite{rauber,sohn_farrar_2001,zhang_zhao_wang_wang_2021,han_mannan_stein_pattipati_bollas_2021,ye_yu_2021} points out that vibration signals provide important information about their mechanical performance. Since the breakage of such an element can lead to a very critical situation, the interest in this field has grown. Many types of cracks can appear in an axle, depending on their shape, direction, and evolution. 

In the field of condition monitoring applied to railway equipment, several studies have been focused on monitoring the condition of track and bearings \cite{silmon_roberts_2010,entezami_roberts_weston_stewart_amini_papaelias_2019}. Other studies analyzed the wheel wear phenomena \cite{article,article2}, and the wheel crack detection \cite{article3}. Traditionally, non-destructive testing techniques were used, based on the complete disassembly of the units to carry out functional tests. However, recent publications are beginning to study the modelling of cracked railway axles with a fault-tolerant approach and crack propagation models \cite{article4,beretta_sangalli_syeda_panggabean_rudlin_2017}. Others, study their dynamic behavior \cite{7920644,rolek_bruni_carboni_2016}, whereas other authors propose to automatically detect cracks \cite{gomez2015,gomez2018,gomez2020,sanchez2020}.

\acf{dl} is a branch of \acf{ml} based on the implementation of \ac{ann} for the extraction and transformation of characteristics based on significant volumes of data. Deep \ac{nn}s are considered hierarchical and multilayered architectures used to extract valuable knowledge from vast amounts of data and use it for classification or regression goals. The crucial features that distinguish \ac{nn}s from any other algorithm are the non-linearity between input and output data and the application of the backpropagation algorithm \cite{pharmaceutics14010183} to optimize the objective function. These ANNs are classified according to the purpose and characteristics of the system. In short, \ac{dl} is a technique that enables algorithms to learn from experience and data, taking advantage of \ac{ann} and multilayer nonlinear processing units. \ac{dl} has contributed to analyzing automatically fault diagnosis and crack classification. In \cite{9327504}, an attention recurrent autoencoder with \ac{lstm}s was proposed to extract time-dependent features from one-dimensional vibration signals. A similar approach was developed in \cite{xu_ma_yan_ma_2021}, where an attention-based-\ac{lstm} and a Random Forest (RF) were used to predict time series trends based on relevant characteristics extracted by wavelet packet analysis. Other authors used auto-encoder models \cite{silva2022supervised} applied to time-frequency spectrograms to locate worn crossings in railway servicing. In other studies, such as \cite{man2022ga}, proposed to use Generative Adversarial Networks (GAN) combined with \ac{LSTM}s to predict the temperature of a high-speed train axle in long-term forecasting. A different approach for automatic failure recognition in photovoltaic systems by thermographic images was proposed in \cite{manno_cipriani_ciulla_di} through the usage of \acf{cnn}s. The application of \ac{cnn}s for identifying failures using vibration sensors was further developed in 
\cite{souza_nascimento_miranda_silva_lepikson_2021} where the aim of alerting when maintenance tasks need to be carried out. From both \cite{9327504,souza_nascimento_miranda_silva_lepikson_2021} we followed the idea of no preprocessing the data with wavelet analysis \cite{xu_ma_yan_ma_2021}. However, we propose not following their approach to dealing with time dependencies, such as LSTMs \cite{9327504, xu_ma_yan_ma_2021}, or 1D-CNN \cite{xu_ma_yan_ma_2021} models. Thus, we propose to use images, that is, a time-frequency representation of the vibration signals.

Therefore, this paper proposes the introduction of a railway axle condition monitoring system based on advanced 2D-CNN architectures applied to time-frequency representations of vibration signals. We introduce a differential CNN structure that captures the statistical properties of the system and enables generalization. This is a critical novelty in the state-of-the-art  and a step forward in condition monitoring systems, improving safety, and reducing costs by extending current inspection intervals. To demonstrate this, several experiments were performed on three Wheelset Assemblies (WA) with different test conditions. In each WA, four types of cracks were tested (no crack, $5.7$mm, $10.9$mm, $15$mm), and classification AUCs of $0.93$, $0.86$, and $0.75$, over each WA respectively, were reported. Thus, improving state-of-the-art methods, such as LSTM \cite{9327504, xu_ma_yan_ma_2021}, 1D-CNNs \cite{souza_nascimento_miranda_silva_lepikson_2021}, and RF \cite{xu_ma_yan_ma_2021}.

The code for reproducing all the experiments proposed in this paper is provided \footnote{\url{https://github.com/antialopezg/Crack-detection-railway-axles-deep-learning}}.

The article is organised as follows: Section \ref{sec:exp} details the experimental system, the conditions under which the tests were performed, and the main characteristics of the vibration signals obtained. Section \ref{sec:model} describes the data preprocessing phase, the model structure, and the selected hyperparameters. In Section \ref{sec:results} the results of our approach and graphical interpretation are provided. Finally, in Section \ref{sec:conclusion} some final remarks are given.

\section{Experimental design} \label{sec:exp}

\begin{figure}[ht]
    \centering
    \includegraphics[width=.7\textwidth]{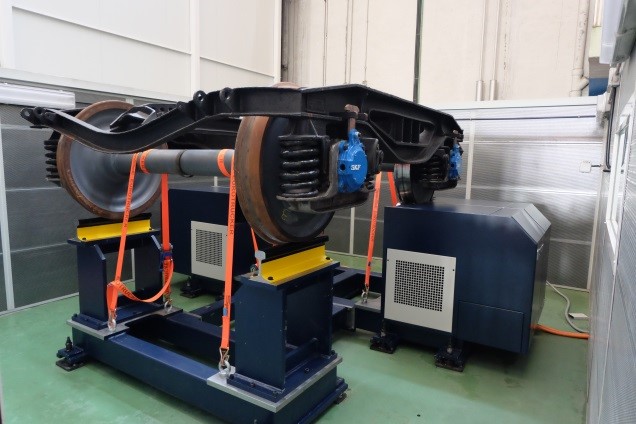}
    \caption[Bogie framework.]{Bogie framework.}
    \label{fig:bogie}
\end{figure}

The railway axle's vibration signals were acquired experimentally, investigating the dynamics of cracked rotors with an experimental system in which full-scaled railway axles were placed \cite{gomez2018,gomez2020}. A bogie test rig was used to build this experimental system. The test rig comprises a fixed bench and a drive system for rolling the axles. It also incorporates a loading system that uses hydraulic actuators to apply a vertical load to the bogie. During the test, the load selected remains constant, replicating real-world situations. The real experimental setup is shown in Fig. \ref{fig:bogie}.

The tests were carried out for 3 WAs, called WA1, WA2, and WA3, thus creating 3 different datasets. For all WAs, the test conditions shown in Table \ref{tab:test_conditions} were tested.

\begin{table}[thb]
    \centering\renewcommand\cellalign{lc}
    \setcellgapes{3pt}\makegapedcells
    \resizebox{\textwidth}{!}{%
    \begin{tabular}{l|c c c c}\hline
         Test condition & \multicolumn{4}{c}{Values} \\
         \hline
         Flaw condition & Healthy shaft (D0) & $5.7$mm (D1) & $10.9$mm (D2) & $15$mm (D3) \\
         Load & $4$t & $10$t &  & \\
         Speed & $20$ km/h & $50$ km/h & &\\
         Wheelsets rotation & Counterclockwise & Clockwise & &\\
         \makecell{Axle box \\accelerometer orientation} & Axial & Lengthwise & Vertical &\\
        \makecell{Axle box \\accelerometer place} & RHS & LHS &  &\\
         \bottomrule
    \end{tabular}}
    \caption{Conditions tested for each WA.}
    \label{tab:test_conditions}
\end{table}

First, the axle was tested without cracks for each of the assemblies tested. Subsequently, without disassembling the axle, three different crack sizes ($5.7$mm, $10.9$mm, and $15$mm, called D1, D2, and D3, respectively) were induced in the central section of the axle. The applied load and speed remained invariant during the tests, replicating natural stationary conditions displayed in Table \ref{tab:test_conditions}. Each axle box had three uniaxial accelerometer sensors oriented in three directions; two radial directions, vertical and horizontal in the direction of motion, and the axial direction in the direction of the axis.  Thus, a total of six accelerometers are set. Furthermore, the rotation of the wheelset was tested in both senses of rotation. 

Vibration signals were acquired using a sampling frequency of $12.8$ kHz and $16384$ points, resulting in signals of duration of $1.28$ s. Then, $58063$ temporal vibration signal samples were obtained and distributed as shown in Table \ref{tab:datadistro}.

\begin{table}[ht]
    \centering
     \resizebox{\textwidth}{!}{%
    \begin{tabular}{c c c c c c}
    \toprule
         Dataset & \# Observations & \# D0 & \# D1 ($5.7$mm) & \# D2 ($10.9$mm) & \# D3 ($15$mm) \\
         \midrule
         WA1 & 12955 & 4762 & 2323 & 2940 & 2930 \\
         WA2 & 20757 & 5220 & 5203 & 5551 & 4783 \\
         WA3 & 11395 & 2996 & 3001 & 3292 & 2106 \\
    \bottomrule
     \end{tabular}}
    \caption{Datasets description. The total number of samples per dataset and their decomposition by each level of defect.}
    \label{tab:datadistro}
\end{table}

\section{Differential convolutional model}
\label{sec:model}

\subsection{Data preprocessing} 
\begin{figure}[ht]
    \centering
    \includegraphics[width=\textwidth]{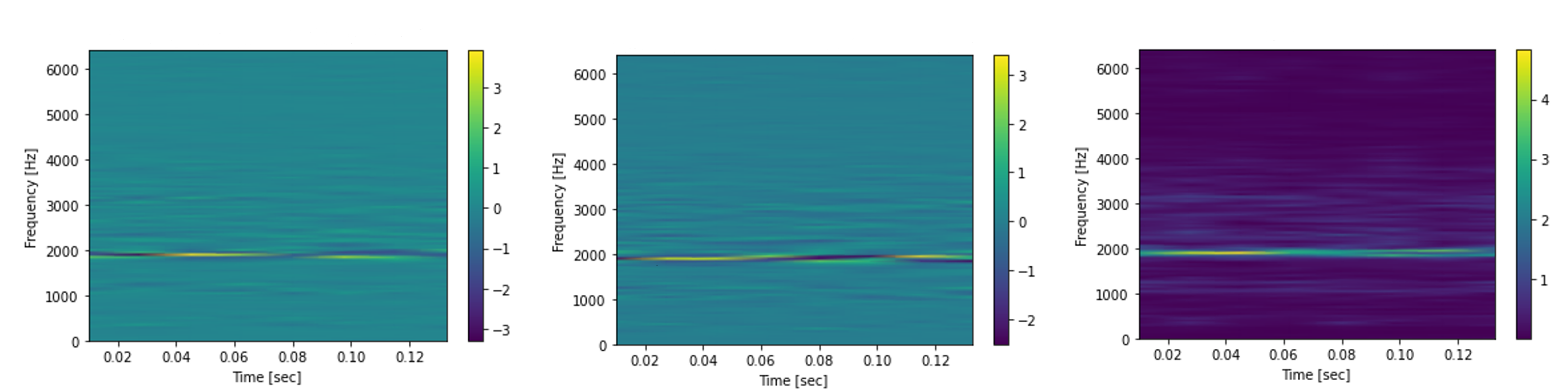}
    \caption{Spectrogram representation of a vibration signal a) real part, b) imaginary part, c) magnitude.}
    \label{fig:spec}
\end{figure}

The vibration signals were reduced to one turn of rotation, subsampled to 2000 points, and normalised. These signals were used in our experiments in two ways: as a raw temporal 1D signal and also converted into the time-frequency domain, via a spectrogram, to explore the 2D spatial correlation. Since the data were transferred to the time-frequency domain, for each vibration signal, a 3-channel image was obtained representing the spectrogram's real, imaginary and magnitude information, as shown in Fig. \ref{fig:spec} for an example.

Then, each of the 4 different defects was categorised, creating a categorical target: no Defect (D0), Defect 1 $5.7mm$ (D1), Defect 2 $10.9$mm (D2), and Defect 3 $15$mm (D3). Data related to the axial direction were discarded as it was found to be not informative in previous studies.

According to the standards, to perform vibration analysis, it is necessary to measure in all three directions (axial, vertical, and longitudinal). However, from what was seen in practice both in this case and in previous studies, such as \cite{gomez2020}, the effects of cracking in this direction are much less visible than in the other directions. This has also been observed from numerical simulations \cite{7920644}. Therefore, only the radial directions of the axle are normally used for this type of system, \cite{beretta_sangalli_syeda_panggabean_rudlin_2017,gomez_ress,gomez2020}. To reduce the computational cost, the axial direction data were excluded.

Once the data was preprocessed and cleaned, it was divided into three partitions: training, validation, and testing. As shown in Table \ref{tab:dataset}, for training, only data corresponding to WA1 was used, specifically $50$\% of the available data. A $30$\% of WA1's data was used for validation. Finally, both the remaining $20$\% of WA1 data and all data acquired from WA2 and WA3 were tested to evaluate whether the model can generalise to a WA never seen.

\begin{table}[]
    \centering
    \begin{tabular}{c c c c c c}
    \toprule
         Training & Validation & & Testing &\\
         \midrule
         WA1 & WA1 & WA1 &  WA2  & WA3 \\
         7255 & 3109 &  2591 & 20757 & 11395 \\ 
         \bottomrule
     \end{tabular}
    \caption{Data used for each step in the models. The first row indicates from which WA is the data while the second row indicates the number of samples used for each purpose.}
    \label{tab:dataset}
\end{table}

\subsection{Model architecture} 

\begin{figure}[htb]
    \centering
    \includegraphics[width=\textwidth]{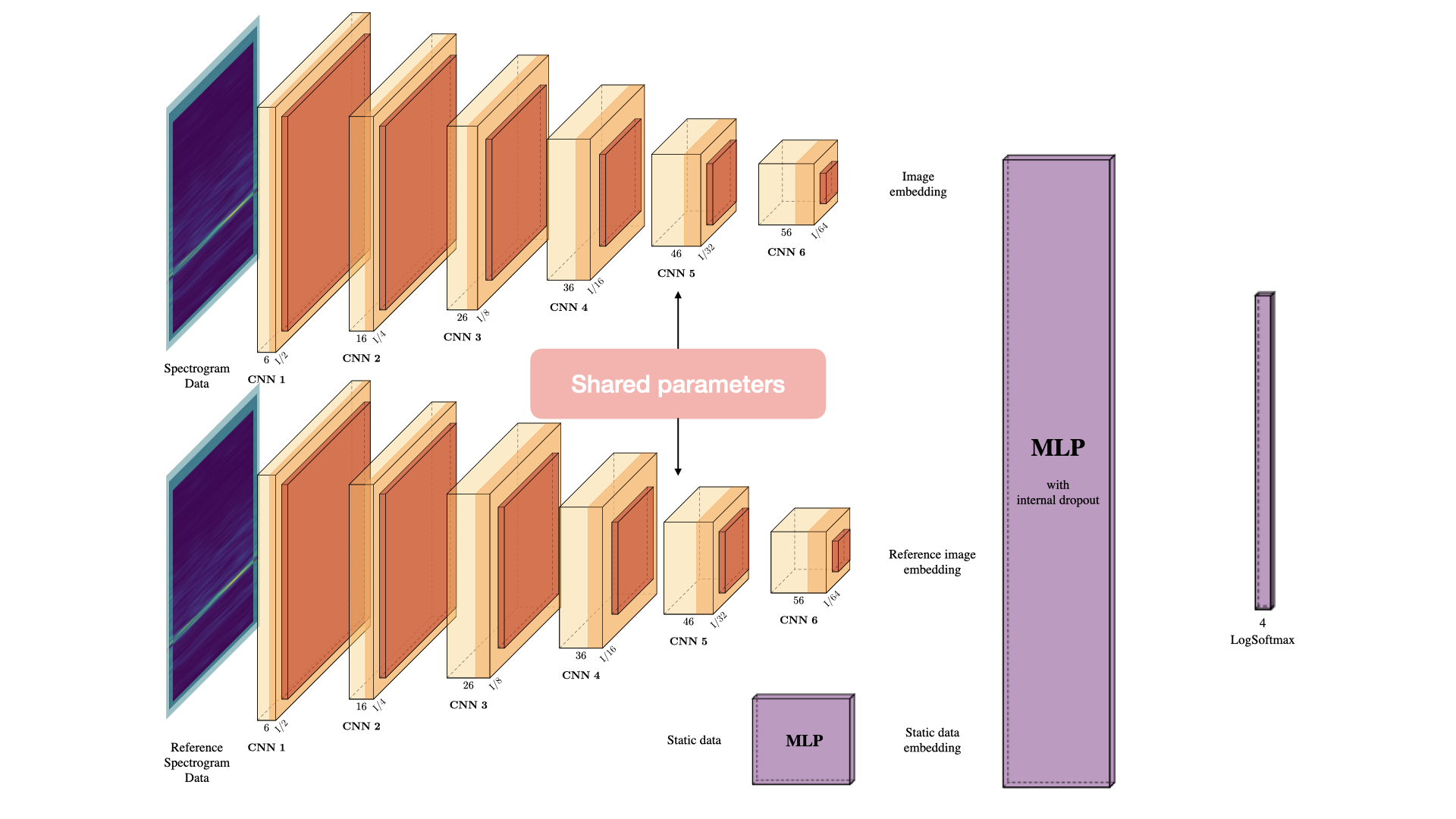}
    \caption{Model architecture. Each convolutional layer follows a $2\times2$ MaxPooling layer, hence, the image is divided by two at each layer. There are three inputs: the spectrograms, the reference spectrograms, and the static data. Thus, three embedding are calculated. Finally, they are all fused to compute the four possible defects: D0, D1, D2, and D3.}
    \label{fig:model}
\end{figure}

The proposed model combined 6 layers of 2D-\ac{cnn} with a \ac{mlp} as a classifier as shown in Fig. \ref{fig:model}. Namely, we used a two-dimensional \ac{cnn} network with 6 convolutional layers of $6,16,26,36,46,56$ channel size with a kernel size of $2$, a stride of $1$, and padding \textit{same}. After the convolutional layers, there was a dropout layer to prevent the network from overfitting. We have not performed exhaustive hyperparameter validation. As such, all of our results can be potentially refined.

To improve generalisation, we have designed a differential approach in which the classifier has two inputs. On the one hand, the spectrogram of the signal to classify. On the other hand, the spectrogram of a reference \textit{healthy} signal axle, is measured at the beginning of the experiment before any axle damage is performed. In this way, we enforce the network to learn the differences between defective and healthy railway axles. Note that the same CNN is used in both branches, i.e., the parameters are shared. That is, feature extraction is common for both signals. Therefore, a fraction of the dataset is left as a reference, i.e., it was essential to randomly extract data corresponding to the healthy shaft so that the system can know the experiments' initial conditions and distinguish between the defects. 

As shown in Fig. \ref{fig:model}, the static characteristics of the signals were processed through an \ac{mlp} with $5$ input features (since we had $5$ static variables, see Table \ref{tab:test_conditions}), and $10$ output features, one for each possible value of the variables. Finally, the vibrations feature vector and the static characteristics feature vector were fused by a fully connected layer with a log-softmax activation function that classified each vibration signal between D0, D1, D2, or D3. Also, a dropout layer is added to prevent overfitting.
 
Through cross-validation, the optimal values for both learning rate and dropout were chosen to be $0.001$ and $0.7$, respectively.

\begin{figure}[htb]
    \centering
    \includegraphics[width=\textwidth]{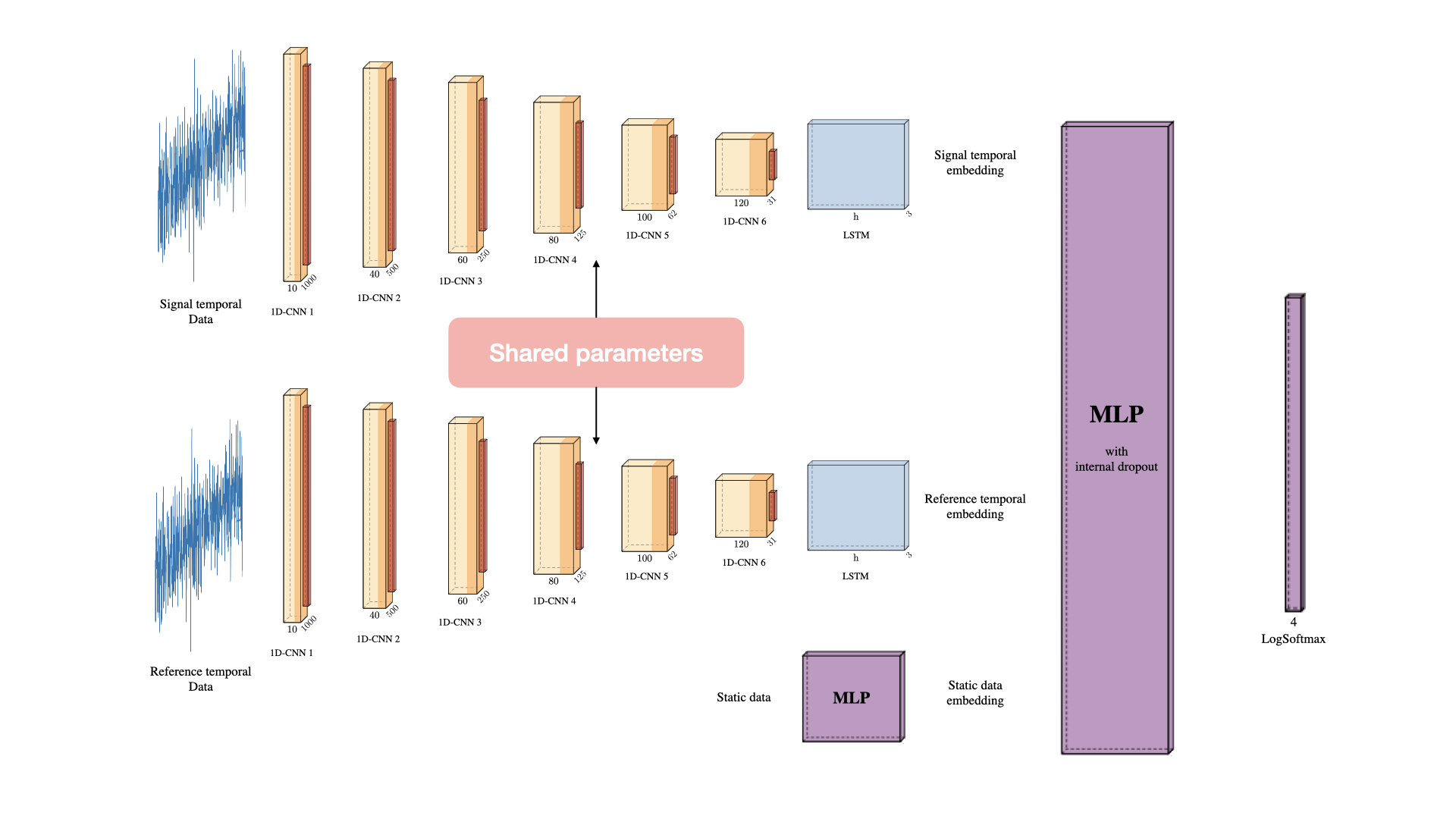}
    \caption{1D-CNN LSTM baseline architecture. There are three inputs: the signals, the reference signals, and the static data. Thus, three embeddings are calculated. Finally, they are all fused to compute the four possible defects: D0, D1, D2, and D3.}
    \label{fig:modelbaseline}
\end{figure}

This article tested different architectures based on different interpretations of the data provided as baselines to compare our proposal. Following the state-of-the-art, we designed and tested other architectures. First, following the literature \cite{9327504, xu_ma_yan_ma_2021} sequential \ac{nn}s were implemented, such as \ac{lstm} \cite{hochreiter1997long}. \ac{lstm}s are a variant of \ac{rnn} that relies on a gated cell to track information through many steps. Therefore, \ac{lstm}s can control the information flow \cite{kalla_2021}. As other authors proposed \cite{souza_nascimento_miranda_silva_lepikson_2021}, a 1D-\ac{cnn}s intended for sequential processing \cite{khandelwal_2020} was also proposed. Furthermore, we also tested the combination of both using a 1D-CNN plus and LSTM where the 1D-CNN is applied to the vibration signal to exploit the time dependency and a unit cell \ac{lstm} was included after the CNN layers to control the information flow over time following the structure shown in Fig. \ref{fig:modelbaseline}.

Moreover, following the literature \cite{xu_ma_yan_ma_2021}, we also used preprocessing packages intended to signal processing such as \textit{tsfresh} \cite{10.1016/j.neucom.2018.03.067} followed by \ac{ml} classifiers. To make a fair comparison with our differential learning, we tested with and without healthy references in all the previous models. Furthermore, we compared our differential learning approach with domain adaptation methods such as \textit{KMM} \cite{de2021adapt} and \textit{TAB} \cite{de2021adapt} followed by \ac{ml} classifiers. These domain adaptation methods are intended to help generalise the models by adapting the new signal seen to the healthy signals of that domain, performing a similar idea to our differential learning approach. In Table \ref{mean} we have compared all these approaches to ours in average performance. To see the detailed results of each baseline, see \ref{appendix}.

\section{Results} \label{sec:results}
Each railway axle vibration signal depends on the conditions tested (place, load, rotation, orientation and speed), which takes two possible values each. Therefore, there are $2^5=32$ possible combinations of these characteristics for which the performance of the models was tested. This performance is evaluated using the \ac{auc}. The \ac{auc} shows the evolution of the false positive ratio versus the true positive ratio or recall, where a 1 indicates a perfect classification of all signals. 

\begin{table}[thb]
\centering
\resizebox{\textwidth}{!}{%
\begin{tabular}{@{}cccccccc@{}}
\toprule
Place & Orientation & Rotation & Load & Speed & \ac{auc}   WA1 & \ac{auc} WA2 & \ac{auc} WA3 \\ 
0 = RHS & 0 = Lengthwise & 0 = Counterclockwise & 0 = 5t & 0 = 20 km/h &  &  &  \\ 
1 = LHS & 1 = Vertical & 1 = Clockwise & 1 = 10t & 1 = 50 km/h &  &  &  \\  \midrule
0    & 0   & 0         & 0    & 0     & 0.845     & 0.896   & 0.797   \\
0    & 0   & 0         & 0    & 1     & 0.954     & 0.943   & 0.699   \\
0    & 0   & 0         & 1    & 0     & 0.75      & 0.917   & 0.692   \\
0    & 0   & 0         & 1    & 1     & 0.962     & \textbf{0.919}   & 0.727   \\
1   & 0   & 0         & 0    & 0     & 0.891     & 0.869   & 0.821   \\
1    & 0   & 0         & 0    & 1     & 0.907     & 0.895   & 0.7     \\
1    & 0   & 0         & 1    & 0     & 0.909     & 0.858   & 0.698   \\
1    & 0   & 0         & 1    & 1     & 0.933     & 0.894   & 0.795   \\
0    & 0   & 1         & 0    & 0     & 0.895     & 0.87    & 0.768   \\
0    & 0   & 1         & 0    & 1     & 0.998     & 0.838   & 0.772   \\
0    & 0   & 1         & 1    & 0     & 0.93      & 0.9     & 0.762   \\
0    & 0   & 1         & 1    & 1     & 0.997     & 0.823   & 0.832   \\
1    & 0   & 1         & 0    & 0     & 0.94      & 0.832   & 0.775   \\
1    & 0   & 1         & 0    & 1     & 0.969     & 0.886   & 0.782   \\
1    & 0   & 1         & 1    & 0     & 0.952     & 0.805   & 0.779   \\
1    & 0   & 1         & 1    & 1     & 1         & 0.793   & 0.669   \\
0    & 1   & 0         & 0    & 0     & 0.885     & 0.923   & 0.741   \\
0    & 1   & 0         & 0    & 1     & 0.985     & 0.917   & 0.682   \\
0    & 1   & 0         & 1    & 0     & 0.752     & 0.927   & 0.688   \\
0    & 1   & 0         & 1    & 1     & 0.914     & 0.855   & 0.785   \\
1    & 1   & 0         & 0    & 0     & 0.96      & 0.906   & 0.806   \\
1    & 1   & 0         & 0    & 1     & 0.973     & 0.823   & 0.629   \\
1    & 1   & 0         & 1    & 0     & 0.915     & 0.901   & 0.702    \\
1    & 1   & 0         & 1    & 1     & 0.972     & 0.794   & 0.764   \\
0    & 1   & 1         & 0    & 0     & 0.904     & 0.872   & 0.754   \\
0 & 1   & 1         & 0    & 1     & 0.992     & 0.814   & 0.778   \\
0    & 1   & 1         & 1    & 0     & 0.916     & 0.809   & 0.789   \\
0    & 1   & 1         & 1    & 1     & 0.994     & 0.841   & 0.704   \\
1    & 1   & 1         & 0    & 0     & 0.919     & 0.795   & 0.743    \\
1    & 1   & 1         & 0    & 1     & 0.995     & 0.77    & \textbf{0.902}   \\
1    & 1   & 1         & 1    & 0     & 0.926     & 0.851   & 0.752   \\
1    & 1   & 1         & 1    & 1     & \textbf{1}         & 0.849   & 0.806   \\ \bottomrule
\end{tabular}%
}
\caption{Results for the classification of the 4 possible defects present in each WA. Each row represents a different test condition scenario. The AUROC columns represent the mean AUROC value scored for the classification of the 4 different defects present in each WA.}
\label{results}
\end{table}

Table \ref{results} shows the \ac{auc} performance for every possible combination of test conditions. As shown, WA1 presented the highest performance, reaching the perfect classification in two configurations. However, WA3 was the most difficult to classify, although no configuration is below 0.66 \ac{auc} performance.

\begin{table}[thb]
\centering
\resizebox{\textwidth}{!}{%
\begin{tabular}{ccccc}
\toprule
Test condition                     & Value & AUROC WA1 & AUROC WA2 & AUROC WA3 \\ \midrule
\multirow{2}{*}{Place}             & RHS     &  $0.91 \pm 0.08$ & $0.88 \pm 0.05$  &   $0.75 \pm 0.05$          \\
                                   & LHS     &$0.95 \pm 0.03$ & $0.85 \pm 0.04$&  $0.76 \pm 0.06$\\ \midrule
\multirow{2}{*}{Orientation}       & Lengthwise     &  $0.93\pm 0.07$ &   $0.87 \pm 0.04$        &   $0.76 \pm 0.05$        \\
                                   & Vertical     &  $0.88 \pm 0.16$ & $0.82 \pm 0.06$          &   $0.76 \pm 0.07$ \\ \midrule
\multirow{2}{*}{Rotation}          & Counterclockwise &  $0.91 \pm 0.07$ & $\mathbf{0.89 \pm 0.04}$          &   $0.73 \pm 0.05$\\
                                   & Clockwise    &  $0.96 \pm 0.04$ & $0.83 \pm 0.04$          &   $\mathbf{0.77 \pm 0.05}$\\\midrule
\multirow{2}{*}{Load}              & 5t    &  $0.94 \pm 0.05$ & $0.87 \pm 0.05$          &   $0.76 \pm 0.06$    \\
                                   & 10t     &  $0.93 \pm 0.08$ & $0.86 \pm 0.05$          &   $0.75 \pm 0.05$\\\midrule
\multirow{2}{*}{Speed}             & 20 km/h     &  $0.89 \pm 0.06$ & $0.87 \pm 0.04$          &   $0.75 \pm 0.04$\\
                                   & 50 km/h     &  $\mathbf{0.97 \pm 0.03}$ & $0.85 \pm 0.05$          &   $0.76 \pm 0.07$  \\  \bottomrule
\end{tabular}%
}
\caption{Summary of Table \ref{results} for each condition. Each value represents the mean and standard deviation w.r.t. AUROC for each value of each condition.}
\label{tab_:summary}
\end{table}

Table \ref{tab_:summary} shows the summary of Table \ref{results} for each condition. When increasing speed, see \textit{Speed 50 km/h} row, it is easier to detect the cracks in the WA1 but it does not improve in the other two WA. Let us remark on the \textit{Counterclockwise} configuration, where it is perfectly generalisable for WA2, achieving similar results as in WA1 but it is the worst performance for WA3. In summary, it is shown that generalising to WA3 is harder in any configuration.

Table \ref{mean} also shows the performance of all baselines compared to the proposed model. As baselines, different approaches were taken. First, the \textit{tsfresh} preprocessing package was used to extract autocorrelated features. Then, it was combined with different classifiers such as Logistic Regressor (LR), RF, or Support Vector Machines (SVM). Then, domain adaptation was proposed to help the generalization. For this purpose, Kernel Mean Matching (KMM) was proposed as an unsupervised domain adaptation method, and TrAdaBoost (TAB) was used as a supervised method. In TAB, healthy railway signals were used as a reference. Further details can be found at \ref{appendix}.  As shown, TAB-LR-Ref performs slightly better in WA3. However, both 1D-CNN-LSTM-Ref and 2D-CNN-Ref outperform all methods, achieving the best overall mean performance. 

\begin{table}[th]
\centering
\begin{tabular}{@{}cccc|c@{}}
\toprule
{Model} & {WA1} & {WA2} & {WA3} & {Overall}\\ \midrule
 \textit{tsfresh}-LR & 0.77 & 0.55 & 0.65 &  0.66\\
  \textit{tsfresh}-SVM & 0.77 & 0.57 & 0.66 &  0.67\\
 \textit{tsfresh}-RF & 0.88 & 0.45 & 0.48 &  0.69\\
 KMM-LR & 0.77 & 0.64 & 0.64 &  0.68\\
KMM-SVM & 0.77 & 0.57 & 0.62 &  0.65\\
 KMM-RF & 0.79 & 0.61 & 0.65 &  0.68\\
  2D-CNN (no reference) & $\mathbf{0.93}$ & 0.56 & 0.46   & 0.65\\
 TAB-LR-Ref & 0.76 & 0.71 & \textbf{0.80} &  0.76\\
TAB-SVM-Ref & 0.77 & 0.69 & 0.71 &  0.72\\
 TAB-RF-Ref & 0.76 & 0.66 & 0.69 &  0.70\\
  1D-CNN-LSTM-Ref & 0.89 & $\mathbf{0.86}$ & 0.79 & \textbf{0.85}\\
\textbf{2D-CNN-Ref} & $\mathbf{0.93}$ & $\mathbf{0.86}$ & 0.75 &  $\mathbf{0.85}$\\ 
 \bottomrule
\end{tabular}%
\caption{Mean \ac{auc} performance for all possible test conditions in each WA of the proposed model and all baselines presented in \ref{appendix}. The \textbf{Ref} label indicates whether the healthy signal of reference was used.}
\label{mean}
\end{table}

\begin{figure}[ht]
    \centering
    \includegraphics[width=0.7\textwidth]{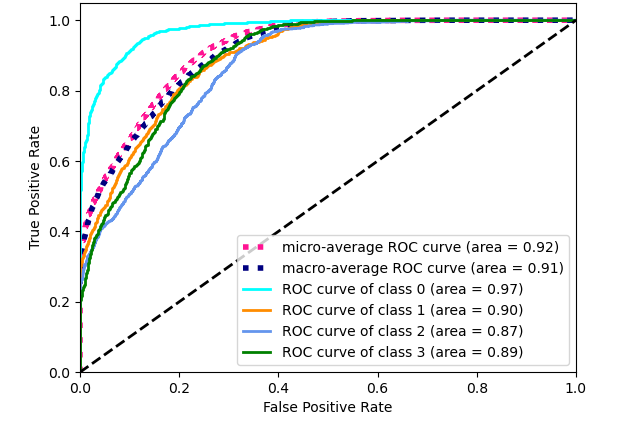}
    \caption{Training ROC curve over WA1.}
    \label{fig:roctr}
\end{figure}

\begin{figure}[ht]
    \centering
    \includegraphics[width=0.7\textwidth]{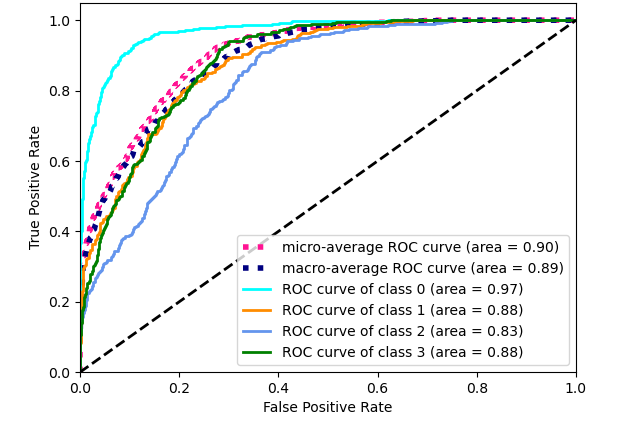}
    \caption{Validation ROC curve over WA1.}
    \label{fig:rocval}
\end{figure}

\begin{figure}[ht]
    \centering
    \includegraphics[width=0.7\textwidth]{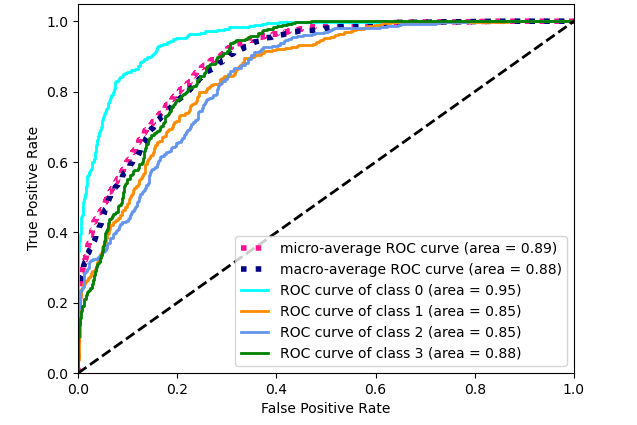}
    \caption{Testing ROC curve over WA1.}
    \label{fig:roctst1}
\end{figure}

\begin{figure}[ht]
    \centering
    \includegraphics[width=0.7\textwidth]{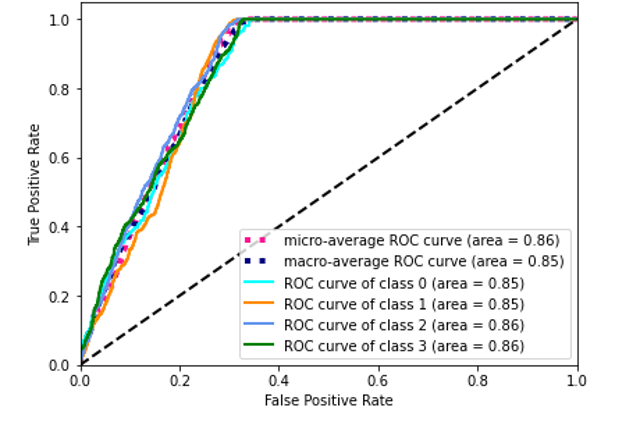}
    \caption{Testing ROC curve over WA2.}
    \label{fig:roctst2}
\end{figure}

\begin{figure}[ht]
    \centering
    \includegraphics[width=0.7\textwidth]{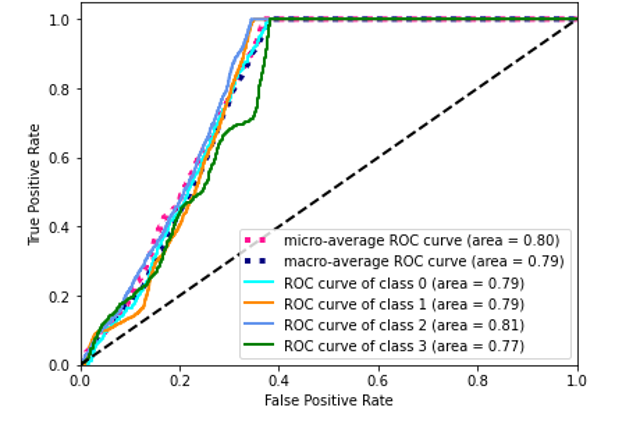}
    \caption{Testing ROC curve over WA3.}
    \label{fig:roctst3}
\end{figure}

Fig. \ref{fig:roctr} and Fig. \ref{fig:rocval} show the ROC curves for training and validation over WA1, respectively. Then, Figs. \ref{fig:roctst1}, \ref{fig:roctst2}, and \ref{fig:roctst3} show testing ROC curves for each WA, classifying the 4 different defects. The proposed model differs between healthy (D0) and defective axles (D1, D2, D3), in all datasets achieving $0.89, 0.85,$ and $0.79$ AUROC scores in each WA, respectively. Furthermore, the system also detects the deepest defect (D3) with an AUROC score of $0.80, 0.85,$ and $0.77$ in each WA, respectively. 


\section{Conclusion} \label{sec:conclusion}

The results concluded that the 1D-CNN-LSTM and 2D-CNN networks are suitable for the diagnosis of cracks in railway axle vibration signals. However, due to the high complexity of dealing with time signals, the 2D-CNN was finally chosen. This solution involves translating the time signals into the time-frequency domain to evaluate spectrograms with image classification methods. 

The results showed that the 2D-CNN architecture outperformed several alternative methods tested. In the first place, preprocessing techniques such as the \textit{tsfresh} \cite{10.1016/j.neucom.2018.03.067} package or domain adaptation packages \cite{de2021adapt} such as KMM or TAB was proven not to be needed. Therefore, 2D-CNN outperformed all baselines and eliminated time-consuming preprocessing steps. Therefore, exploiting the spatial correlation of the spectrograms of the railway axle vibration signals allows the diagnosis of railway axle cracks. Second, it was proved necessary to include the reference information on the WA in which the vibration signals were generated to generalise to any setup. In this way, information was passed on about the state of the healthy axle, allowing the system to distinguish healthy behaviour from defective one.

In future work, we will analyze which regions of the spectrogram influence the classifier's decision by including attention models, also known as Transformers \cite{vaswani2017attention}.

\section*{Acknowledgments}

This work has been developed within the project \emph{Sistema de monitorización de estado para detección de fisuras en ejes ferroviarios} (SMEPDFEF-CM-UC3M), 2021 Call for grants to carry out interdisciplinary R\&D projects for young doctors of the Carlos III University from Madrid. The work was also supported by MCIN AEI [grant numbers RTI2018-099655-B-100 and PID2021-123182OB-I00]; by the Comunidad de Madrid [grant numbers IND2022/TIC-23550]; by the European Union (FEDER and the European Research Council (ERC) through the European Unions Horizon 2020 research innovation program [grant number 714161]; and by the IISGM [grant INTRAMURAL]. 

\bibliography{paper-principal}

\newpage
\appendix

\section{Additional experiments}
\label{appendix}

\setcounter{table}{0}
\renewcommand{\thetable}{A.1.\arabic{table}}

Other architectures were designed to test the effectiveness and reliability of the results obtained with the 2D-\ac{cnn}s:

\begin{itemize}

    \item \textbf{Not reference-based evaluation}: the input into 2D-CNN was only the spectrogram to be classified without the reference of a healthy axle.
    
    \item \textbf{1DCNN + LSTM}: the vibration signal was used in the time domain, that is, no spectrogram was used. A 1D-CNN is applied to the vibration signal to exploit the time dependency. A \ac{lstm} unit cell was included after the CNN layers to control the information flow over time. 
    
    \item \textbf{\textit{Tsfresh} \cite{10.1016/j.neucom.2018.03.067} feature extraction and \ac{ml} classifier}: the \textit{tsfresh} library was used to extract features from temporal signals and reduce dimensionality. These features were used as input to different classifiers such as an LR, an SVM, or an RF.

    \item \textbf{\textit{KMM} \cite{de2021adapt} unsupervised domain adaptation and \ac{ml} classifier}: the unsupervised KMM method was used to adapt the domain using reference healthy signals unlabeled. This new domain was used as input to different classifiers such as an LR, an SVM, or an RF.

    \item \textbf{\textit{TAB} \cite{de2021adapt} supervised domain adaptation and \ac{ml} classifier}: the supervised KMM method was used to adapt the domain using reference healthy signals labelled. This new domain was used as input to different classifiers such as an LR, an SVM, or an RF.
\end{itemize}

\subsection{Results}

Results by configuration and dataset are displayed in terms of \ac{auc} for each of the three approaches.

\subsubsection{Not references}

\begin{table}[ht]
\resizebox{\textwidth}{!}{%
\begin{tabular}{@{}cccccccc@{}}
\toprule
Place & Orientation & Rotation & Load & Speed & \ac{auc}   WA1 & \ac{auc} WA2 & \ac{auc} WA3 \\ 
0 = RHS & 0 = Lengthwise & 0 = Counterclockwise & 0 = 5t & 0 = 20 km/h &  &  &  \\ 
1 = LHS & 1 = Vertical & 1 = Clockwise & 1 = 10t & 1 = 50 km/h &  &  &  \\  \midrule
0.0   & 0.0         & 0.0      & 0.0  & 0.0   & 0.867   & 0.682   & 0.676     \\
0.0   & 0.0         & 0.0      & 0.0  & 1.0   & 0.978   & 0.673   & 0.42     \\
0.0   & 0.0         & 0.0      & 1.0  & 0.0   & 0.836   & 0.781   & 0.326    \\
0.0   & 0.0         & 0.0      & 1.0  & 1.0   & 0.977   & 0.776   & 0.194    \\
1.0   & 0.0         & 0.0      & 0.0  & 0.0   & 0.902   & 0.625   & 0.498     \\
1.0   & 0.0         & 0.0      & 0.0  & 1.0   & 0.992   & 0.447   & 0.468    \\
1.0   & 0.0         & 0.0      & 1.0  & 0.0   & 0.819   & 0.7     & 0.29      \\
1.0   & 0.0         & 0.0      & 1.0  & 1.0   & 0.987   & 0.549   & 0.679   \\
0.0   & 0.0         & 1.0      & 0.0  & 0.0   & 0.928   & 0.65    & 0.47    \\
0.0   & 0.0         & 1.0      & 0.0  & 1.0   & 1.0     & 0.438   & 0.417       \\
0.0   & 0.0         & 1.0      & 1.0  & 0.0   & 0.925   & 0.652   & 0.489      \\
0.0   & 0.0         & 1.0      & 1.0  & 1.0   & 0.998   & 0.445   & 0.664    \\
1.0   & 0.0         & 1.0      & 0.0  & 0.0   & 0.907   & 0.442   & 0.528     \\
1.0   & 0.0         & 1.0      & 0.0  & 1.0   & 0.996   & 0.349   & 0.757     \\
1.0   & 0.0         & 1.0      & 1.0  & 0.0   & 0.87    & 0.439   & 0.487     \\
1.0   & 0.0         & 1.0      & 1.0  & 1.0   & 0.999   & 0.598   & 0.597   \\
0.0   & 1.0         & 0.0      & 0.0  & 0.0   & 0.864   & 0.64    & 0.354    \\
0.0   & 1.0         & 0.0      & 0.0  & 1.0   & 0.999   & 0.392   & 0.493   \\
0.0   & 1.0         & 0.0      & 1.0  & 0.0   & 0.831   & 0.639   & 0.226     \\
0.0   & 1.0         & 0.0      & 1.0  & 1.0   & 0.957   & 0.457   & 0.398    \\
1.0   & 1.0         & 0.0      & 0.0  & 0.0   & 0.895   & 0.648   & 0.583    \\
1.0   & 1.0         & 0.0      & 0.0  & 1.0   & 1.0     & 0.502   & 0.498      \\
1.0   & 1.0         & 0.0      & 1.0  & 0.0   & 0.835   & 0.667   & 0.219      \\
1.0   & 1.0         & 0.0      & 1.0  & 1.0   & 0.995   & 0.55    & 0.446     \\
0.0   & 1.0         & 1.0      & 0.0  & 0.0   & 0.877   & 0.465   & 0.378     \\
0.0   & 1.0         & 1.0      & 0.0  & 1.0   & 1.0     & 0.381   & 0.186      \\
0.0   & 1.0         & 1.0      & 1.0  & 0.0   & 0.89    & 0.591   & 0.491     \\
0.0   & 1.0         & 1.0      & 1.0  & 1.0   & 0.99    & 0.643   & 0.563     \\
1.0   & 1.0         & 1.0      & 0.0  & 0.0   & 0.931   & 0.657   & 0.51      \\
1.0   & 1.0         & 1.0      & 0.0  & 1.0   & 0.998   & 0.285   & 0.729      \\
1.0   & 1.0         & 1.0      & 1.0  & 0.0   & 0.86    & 0.62    & 0.491    \\
1.0   & 1.0         & 1.0      & 1.0  & 1.0   & 0.999   & 0.521   & 0.334      \\ \bottomrule
\end{tabular}%
}

\caption{\ac{auc} results non referenced-based model.}
\label{tab:nonref}
\end{table}

As shown in Table \ref{tab:nonref}, the mean \ac{auc} for WA1 was $0.93$, as in the proposal of the main document. However, the mean \ac{auc} was $0.56$ and $0.46$ for WA2 and WA3, respectively, which implies that if we omitted the reference spectrograms in the \ac{nn}, it was not able to generalise to different WA. In other words, the system needs information about the healthy axle to distinguish the initial configuration and thus identify defects.

\subsubsection{1D-CNN-LSTM}

\begin{table}[ht]
\resizebox{\textwidth}{!}{%
\begin{tabular}{@{}cccccccc@{}}
\toprule
Place & Orientation & Rotation & Load & Speed & \ac{auc}   WA1 & \ac{auc} WA2 & \ac{auc} WA3 \\ 
0 = RHS & 0 = Lengthwise & 0 = Counterclockwise & 0 = 5t & 0 = 20 km/h &  &  &  \\ 
1 = LHS & 1 = Vertical & 1 = Clockwise & 1 = 10t & 1 = 50 km/h &  &  &  \\  \midrule
0.0   & 0.0         & 0.0      & 0.0  & 0.0   & 0.818   & 0.907   & 0.769     \\
0.0   & 0.0         & 0.0      & 0.0  & 1.0   & 0.887   & 0.884   & 0.715     \\
0.0   & 0.0         & 0.0      & 1.0  & 0.0   & 0.765   & 0.889   & 0.84      \\
0.0   & 0.0         & 0.0      & 1.0  & 1.0   & 0.893   & 0.913   & 0.718     \\
1.0   & 0.0         & 0.0      & 0.0  & 0.0   & 0.86    & 0.87    & 0.882     \\
1.0   & 0.0         & 0.0      & 0.0  & 1.0   & 0.889   & 0.864   & 0.634   \\
1.0   & 0.0         & 0.0      & 1.0  & 0.0   & 0.907   & 0.89    & 0.863    \\
1.0   & 0.0         & 0.0      & 1.0  & 1.0   & 0.869   & 0.9     & 0.841  \\
0.0   & 0.0         & 1.0      & 0.0  & 0.0   & 0.882   & 0.902   & 0.765      \\
0.0   & 0.0         & 1.0      & 0.0  & 1.0   & 0.954   & 0.834   & 0.817     \\
0.0   & 0.0         & 1.0      & 1.0  & 0.0   & 0.925   & 0.85    & 0.81   \\
0.0   & 0.0         & 1.0      & 1.0  & 1.0   & 0.937   & 0.844   & 0.814    \\
1.0   & 0.0         & 1.0      & 0.0  & 0.0   & 0.922   & 0.858   & 0.828    \\
1.0   & 0.0         & 1.0      & 0.0  & 1.0   & 0.942   & 0.841   & 0.833     \\
1.0   & 0.0         & 1.0      & 1.0  & 0.0   & 0.908   & 0.805   & 0.785   \\
1.0   & 0.0         & 1.0      & 1.0  & 1.0   & 0.897   & 0.796   & 0.762      \\
0.0   & 1.0         & 0.0      & 0.0  & 0.0   & 0.859   & 0.898   & 0.797     \\
0.0   & 1.0         & 0.0      & 0.0  & 1.0   & 0.876   & 0.895   & 0.728   \\
0.0   & 1.0         & 0.0      & 1.0  & 0.0   & 0.799   & 0.887   & 0.869     \\
0.0   & 1.0         & 0.0      & 1.0  & 1.0   & 0.804   & 0.882   & 0.801     \\
1.0   & 1.0         & 0.0      & 0.0  & 0.0   & 0.971   & 0.917   & 0.824     \\
1.0   & 1.0         & 0.0      & 0.0  & 1.0   & 0.888   & 0.932   & 0.702     \\
1.0   & 1.0         & 0.0      & 1.0  & 0.0   & 0.933   & 0.883   & 0.868      \\
1.0   & 1.0         & 0.0      & 1.0  & 1.0   & 0.873   & 0.925   & 0.668     \\
0.0   & 1.0         & 1.0      & 0.0  & 0.0   & 0.896   & 0.827   & 0.75     \\
0.0   & 1.0         & 1.0      & 0.0  & 1.0   & 0.925   & 0.866   & 0.79       \\
0.0   & 1.0         & 1.0      & 1.0  & 0.0   & 0.885   & 0.816   & 0.818   \\
0.0   & 1.0         & 1.0      & 1.0  & 1.0   & 0.986   & 0.825   & 0.75      \\
1.0   & 1.0         & 1.0      & 0.0  & 0.0   & 0.916   & 0.881   & 0.781    \\
1.0   & 1.0         & 1.0      & 0.0  & 1.0   & 0.916   & 0.881   & 0.855     \\
1.0   & 1.0         & 1.0      & 1.0  & 0.0   & 0.873   & 0.786   & 0.836     \\
1.0   & 1.0         & 1.0      & 1.0  & 1.0   & 0.904   & 0.81    & 0.809    \\ \bottomrule
\end{tabular}%
}
\caption{\ac{auc} results 1D-CNN-LSTM model.}
\label{tab:1dcnnlstm}
\end{table}

Table \ref{tab:1dcnnlstm} shows a similar performance to the main proposal presented in the article. The mean \ac{auc} was $0.89, 0.86,$ and $0.79$ for WA1, WA2 and WA3, respectively.  However, the overall mean \ac{auc} for the 1D-CNN-LSTM case was $0.855$, while the 2D-CNN was $0.867$. Therefore, 2D-CNN is more generalised.

In addition, images are easy to handle with a \ac{cnn} that exploits spatial correlation.

\subsubsection{Feature extraction and a Machine Learning classifier}

\begin{table}[ht]
\resizebox{\textwidth}{!}{%
\begin{tabular}{@{}cccccccc@{}}
\toprule
Place & Orientation & Rotation & Load & Speed & \ac{auc}   WA1 & \ac{auc} WA2 & \ac{auc} WA3 \\ 
0 = RHS & 0 = Lengthwise & 0 = Counterclockwise & 0 = 5t & 0 = 20 km/h &  &  &  \\ 
1 = LHS & 1 = Vertical & 1 = Clockwise & 1 = 10t & 1 = 50 km/h &  &  &  \\  \midrule
0.0   & 0.0         & 0.0      & 0.0  & 0.0   & 0.892   & 0.365   & 0.506    \\
0.0   & 0.0         & 0.0      & 0.0  & 1.0   & 0.899   & 0.407   & 0.535    \\
0.0   & 0.0         & 0.0      & 1.0  & 0.0   & 0.875   & 0.46    & 0.505     \\
0.0   & 0.0         & 0.0      & 1.0  & 1.0   & 0.874   & 0.451   & 0.526      \\
1.0   & 0.0         & 0.0      & 0.0  & 0.0   & 0.855   & 0.436   & 0.493    \\
1.0   & 0.0         & 0.0      & 0.0  & 1.0   & 0.885   & 0.498   & 0.513\\
1.0   & 0.0         & 0.0      & 1.0  & 0.0   & 0.87    & 0.437   & 0.474    \\
1.0   & 0.0         & 0.0      & 1.0  & 1.0   & 0.898   & 0.491   & 0.481       \\
0.0   & 0.0         & 1.0      & 0.0  & 0.0   & 0.835   & 0.452   & 0.416      \\
0.0   & 0.0         & 1.0      & 0.0  & 1.0   & 0.913   & 0.414   & 0.478    \\
0.0   & 0.0         & 1.0      & 1.0  & 0.0   & 0.936   & 0.451   & 0.467    \\
0.0   & 0.0         & 1.0      & 1.0  & 1.0   & 0.872   & 0.441   & 0.451    \\
1.0   & 0.0         & 1.0      & 0.0  & 0.0   & 0.875   & 0.461   & 0.496      \\
1.0   & 0.0         & 1.0      & 0.0  & 1.0   & 0.858   & 0.489   & 0.486     \\
1.0   & 0.0         & 1.0      & 1.0  & 0.0   & 0.858   & 0.423   & 0.465    \\
1.0   & 0.0         & 1.0      & 1.0  & 1.0   & 0.88    & 0.456   & 0.486     \\
0.0   & 1.0         & 0.0      & 0.0  & 0.0   & 0.86    & 0.41    & 0.515      \\
0.0   & 1.0         & 0.0      & 0.0  & 1.0   & 0.884   & 0.442   & 0.434    \\
0.0   & 1.0         & 0.0      & 1.0  & 0.0   & 0.856   & 0.441   & 0.486     \\
0.0   & 1.0         & 0.0      & 1.0  & 1.0   & 0.903   & 0.473   & 0.494     \\
1.0   & 1.0         & 0.0      & 0.0  & 0.0   & 0.915   & 0.454   & 0.438     \\
1.0   & 1.0         & 0.0      & 0.0  & 1.0   & 0.908   & 0.471   & 0.444     \\
1.0   & 1.0         & 0.0      & 1.0  & 0.0   & 0.89    & 0.476   & 0.494    \\
1.0   & 1.0         & 0.0      & 1.0  & 1.0   & 0.929   & 0.448   & 0.476     \\
0.0   & 1.0         & 1.0      & 0.0  & 0.0   & 0.906   & 0.48    & 0.545     \\
0.0   & 1.0         & 1.0      & 0.0  & 1.0   & 0.923   & 0.464   & 0.49     \\
0.0   & 1.0         & 1.0      & 1.0  & 0.0   & 0.895   & 0.451   & 0.498     \\
0.0   & 1.0         & 1.0      & 1.0  & 1.0   & 0.837   & 0.433   & 0.5        \\
1.0   & 1.0         & 1.0      & 0.0  & 0.0   & 0.873   & 0.478   & 0.526     \\
1.0   & 1.0         & 1.0      & 0.0  & 1.0   & 0.872   & 0.492   & 0.457      \\
1.0   & 1.0         & 1.0      & 1.0  & 0.0   & 0.853   & 0.457   & 0.449     \\
1.0   & 1.0         & 1.0      & 1.0  & 1.0   & 0.827   & 0.465   & 0.526      \\ \bottomrule
\end{tabular}%
}
\caption{\ac{auc} results Random Forest model.}
\label{tab:rf}
\end{table}

The last experiment consisted of two phases. The first phase, the feature extraction step, was based on the use of the \textit{tsfresh} library to extract and select relevant features from the signals and use them as input to the classifier instead of the vibration signal. The second phase consisted of performing cross-validation to train a Random Forest classifier and evaluate this model in each possible configuration. 

As seen in Table \ref{tab:rf}, this model does not generalise to the rest of the datasets, so there was a lot of overfitting. The mean \ac{auc} per dataset was much lower than in previous cases, being 0.88, 0.45, and 0.48 for WA1, WA2 and WA3, respectively.

\end{document}